\pdfoutput=1
\documentclass[11pt]{article}

\PassOptionsToPackage{table}{xcolor}
\usepackage[preprint]{acl}

\usepackage{times}
\usepackage{latexsym}
\usepackage[T1]{fontenc}
\usepackage[utf8]{inputenc}
\usepackage{microtype}
\usepackage{inconsolata}

\usepackage{booktabs}
\usepackage{amsfonts}
\usepackage{amsmath}
\usepackage{amssymb}
\usepackage{graphicx}
\usepackage{algorithm}
\usepackage{algorithmic}
\usepackage{multirow}
\usepackage{subcaption}
\usepackage{enumitem}
\usepackage{xcolor}
\usepackage{colortbl}
\usepackage{arydshln}

\newcommand{\method}{CoverPruner}

\title{Who Speaks for the Pruned? \\
Visual Token Pruning as Coverage Optimization}

\author{
  Qingchan Zhu\textsuperscript{1}\Thanks{Equal contribution.},
  Weihang You\textsuperscript{1}\footnotemark[1],
  Hanqi Jiang\textsuperscript{1},
  Changdi Yang\textsuperscript{2},
  Tianming Liu\textsuperscript{1},
  Geng Yuan\textsuperscript{1}\Thanks{Corresponding Author.} \\
  \textsuperscript{1}School of Computing, University of Georgia \\
  \textsuperscript{2}College of Engineering, Northeastern University \\
  \texttt{\{qingchan.zhu, weihang.you, hanqi.jiang, tliu, geng.yuan\}@uga.edu} \\
  \texttt{yang.changd@northeastern.edu}
}

\begin{document}

\maketitle

\begin{abstract}
Visual token pruning reduces the inference cost of vision-language models (VLMs), but most methods only ask which tokens to keep. This retained-token view can keep redundant high-scoring tokens while leaving discarded evidence without a close representative. We propose \method{}, a training-free pruner that asks the complementary demand-side question: after a token is removed, which surviving original token represents it for the target VLM? \method{} formulates pruning as Representational Coverage Maximization (RCM), covering the full projected visual-token set with query-weighted demand. It instantiates RCM with projector-space coverage and a lightweight first-layer attention probe. Across multiple VLM architectures and compression rates, \method{} achieves the best average accuracy among all compared methods, with the largest gains usually appearing under aggressive compression.
\end{abstract}

\section{Introduction}
\label{sec:intro}

Vision-language models (VLMs)~\cite{liu2023llava, liu2024improved, li2024llavaonevision, bai2025qwen25vl} extend language models to visual inputs by encoding images into sequences of visual tokens. Modern systems produce hundreds of tokens per image at standard resolution and several thousand at high resolution~\cite{liu2024llavanext}, with video inputs reaching tens of thousands~\cite{zhang2024videoinstructiontuning}. Attention scales quadratically with sequence length, so the volume of visual tokens dominates inference cost. Visual token pruning addresses this by retaining only a small subset of visual tokens during inference, but a ranker may keep redundant salient patches while dropping a small sign, diagram label, or contextual region needed elsewhere.

\begin{figure*}[t]
    \centering
    \resizebox{0.98\textwidth}{!}{%
        \includegraphics{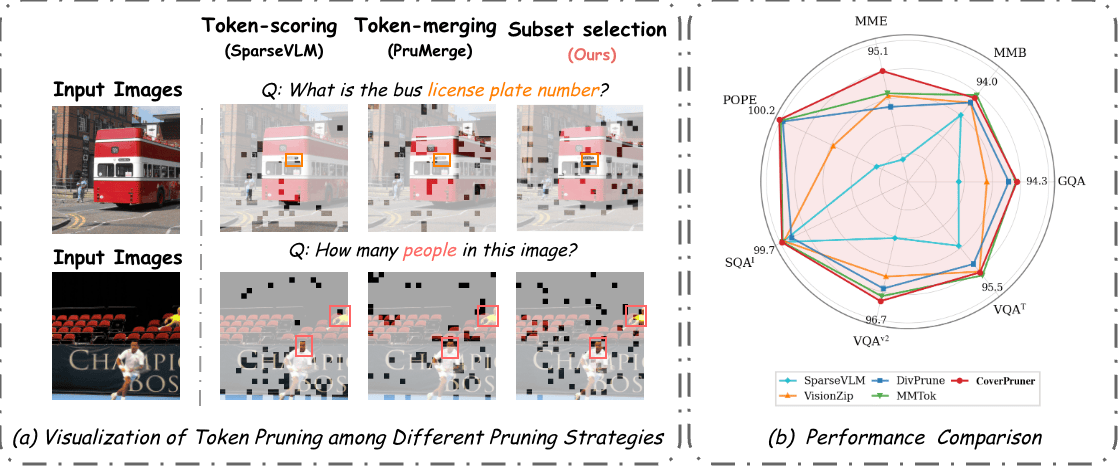}%
    }
    \caption{\textbf{Comparison of visual token pruning methods on LLaVA-1.5-7B.} \textbf{(a)} Token selection paradigms: scoring keeps individually salient tokens, merging fuses similar tokens into new visual tokens, and \method{} keeps original tokens that cover the visual evidence removed by pruning. \textbf{(b)} Performance retention (Rel.\% of full-model accuracy) across benchmarks at 64 retained tokens.}
    \label{fig:intro}
\end{figure*}

Prior work typically defines pruning by what should be kept. Token-scoring methods~\cite{chen2024fastv, xing2024pyramiddrop, zhang2024sparsevlm, yang2024visionzip, wen2025dart, li2026why} rank tokens independently using attention or feature-based importance; for example, they may keep several high-scoring tokens from the same salient object while leaving a lower-scored text patch or contextual region without a representative. Token-merging methods~\cite{shang2025prumerge, song2024trim} reduce redundancy by fusing similar visual tokens, but the fused representation can dilute fine-grained evidence from an individual patch, such as small text, object boundaries, or chart marks. Set-level methods~\cite{alvar2025divprune, dong2026mmtok, guo2025crop} introduce diversity, coverage, or query-guided localization, but they do not define coverage over the target VLM's projected visual-token space. Together, these methods improve which tokens are scored, fused, or selected, but they do not directly model the demand-side question: pruning is not only deciding which tokens survive, but deciding who speaks for the tokens that disappear.

This demand-side view suggests a simple criterion: a single sky token can cover a redundant sky region, while distinct text patches on a street sign demand separate representatives. We therefore formulate visual token pruning as Representational Coverage Maximization (RCM): the full visual-token set defines the demand to be covered, the retained subset supplies representatives, and each demand token is weighted by query relevance.

\method{} turns the RCM objective into a training-free VLM pruner. It measures coverage in the projector output space, where visual tokens are passed to the language model; calibrates similarities by mean-centering and clipping them; and estimates query relevance with a lightweight first-layer attention probe. The resulting selector is instruction-conditioned, keeps the original representations of retained tokens, and is used unchanged across models, benchmarks, and token budgets in our experiments.

Experiments across VLM architectures and compression rates show that \method{} achieves the best average accuracy among all compared methods, with especially clear gains at aggressive token budgets (Figure~\ref{fig:intro}(b)).

\noindent Our main contributions are:
\begin{itemize}[leftmargin=*, itemsep=0pt, parsep=0pt, topsep=2pt]
    \item We reframe visual token pruning as RCM, shifting the criterion from survivor importance to demand coverage in the target VLM's representation space.
    \item We propose \method{}, a training-free pruner that realizes this demand-side view for VLMs, selecting original projected tokens for the current instruction without external localization.
    \item Experiments across multiple VLMs and compression rates show strong average gains, and find that relevance is more effective for weighting coverage than ranking tokens.
\end{itemize}

\section{Related Work}
\label{sec:related}

\paragraph{Visual token pruning in VLMs.}
The growing cost of visual tokens in VLMs has motivated a broad line of token reduction methods. Pre-fusion compression methods~\cite{li2024llamavid, cai2024matryoshka} reduce visual tokens through architectural changes and additional training, which is orthogonal to our training-free setting. Inference-time pruning methods instead remove tokens during the forward pass, typically using attention signals~\cite{chen2024fastv, xing2024pyramiddrop, zhang2024sparsevlm, ye2025fitpruner} or visual-feature redundancy~\cite{wen2025dart, yang2024visionzip, jeddi2025similarity} to decide which tokens to retain. Token-merging methods~\cite{shang2025prumerge, song2024trim} further reduce redundancy by fusing similar tokens. These methods establish that visual token sequences are highly compressible, but most decisions are still driven by token-local scores, local redundancy, or modified token representations rather than by an explicit representation requirement for the discarded tokens.

Recent methods also combine instruction-conditioned diversity (CDPruner~\cite{zhang2025cdpruner}), projection-layer zeroth-order sensitivity with diversity (ZOO-Prune~\cite{kim2026zooprune}), and image-adaptive attention--diversity selection (AgilePruner~\cite{baek2026agilepruner}).

\paragraph{Representative selection and coverage.}
Our work is more closely related to methods that treat pruning as representative selection. DivPrune~\cite{alvar2025divprune} selects a diverse subset, MMTok~\cite{dong2026mmtok} formulates multimodal coverage in a shared encoder space, and CROP~\cite{guo2025crop} uses query-conditioned contextual-region localization. \method{} differs in what is represented: retained original projected tokens cover the full projected token set in the consuming VLM's own input space. This connects visual token pruning to representative subset selection objectives studied in coreset construction~\cite{mirzasoleiman2020coresets, wei2014submodular} and active learning~\cite{sener2018active}, while adapting them to query-conditioned VLM inference. 

\section{Method}
\label{sec:method}

\begin{figure*}[t]
    \centering
    \resizebox{\textwidth}{!}{%
        \includegraphics{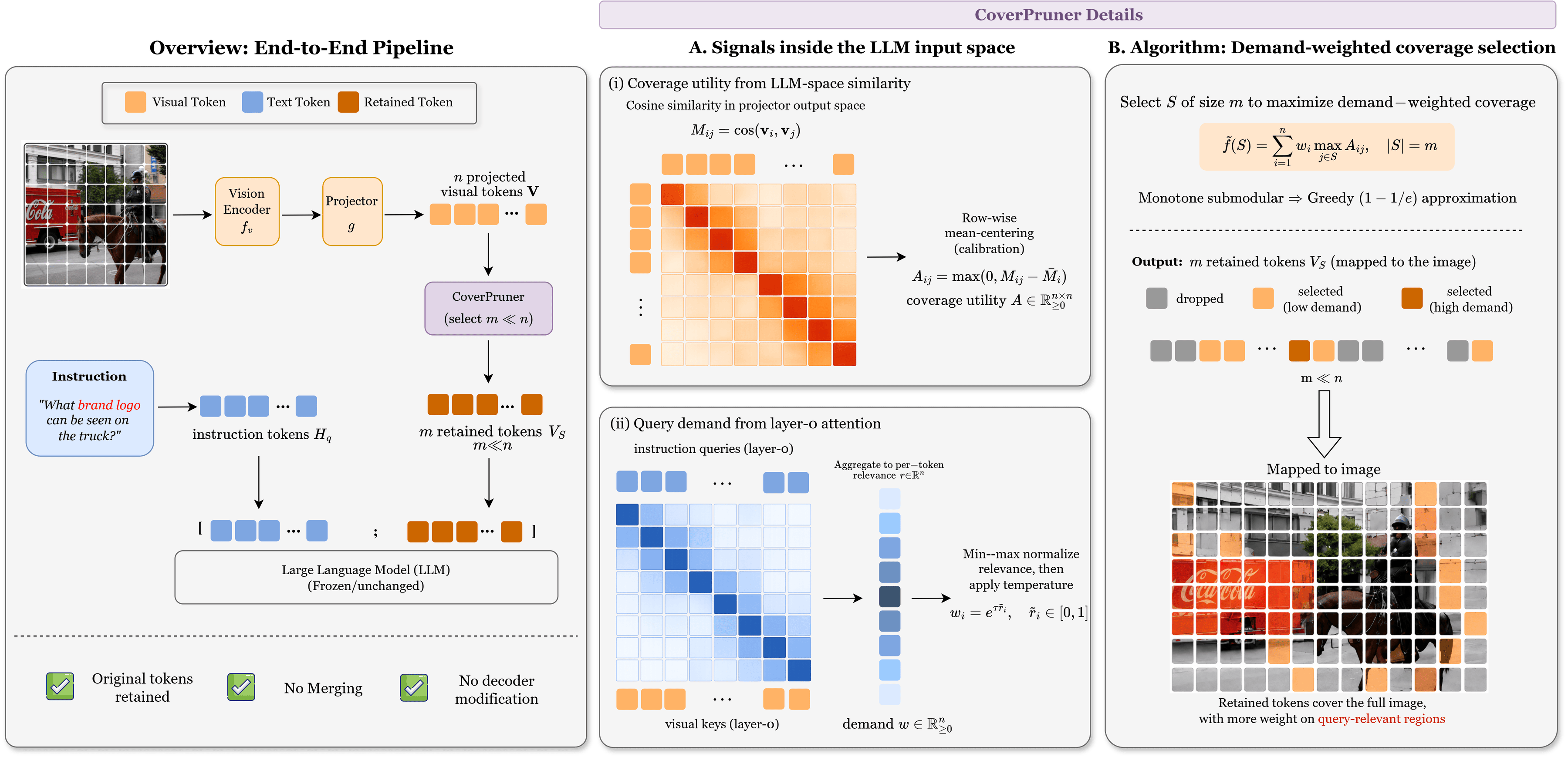}%
    }
    \caption{\textbf{Schematic overview of \method{}.} The left panel shows the end-to-end pipeline; the right panels detail (A) coverage and query-demand signal construction in the LLM input space and (B) demand-weighted coverage selection. Token grids are shown as visual abstractions.}
    \label{fig:framework}
\end{figure*}

\subsection{Design Overview}

\method{} operates at the interface where visual evidence enters the language model. Given an image $X_v$, the vision encoder $f_v$ and projector $g$ produce projected visual tokens $\mathbf{V}=g(f_v(X_v))\in\mathbb{R}^{n\times d}$, which are concatenated with instruction tokens $\mathbf{H}_q$ and consumed by the decoder $f_\phi$. For a token budget $m\ll n$, \method{} keeps an ordered subset $\mathbf{V}_S$ ($|S|=m$) and feeds $[\mathbf{V}_S;\mathbf{H}_q]$ to the unchanged decoder, without merging tokens or altering their representations.

Figure~\ref{fig:framework} shows the end-to-end pipeline on the left and the signal-construction and selection details on the right. \method{} compares projected visual tokens in the decoder input space, converts the similarities into non-negative coverage utilities, estimates instruction-conditioned demand with a lightweight layer-0 attention probe, and selects retained tokens by marginal coverage gain. The next subsections formalize the coverage objective, LLM-anchored similarity and relevance, and demand-weighted coverage selection under a fixed budget.

\subsection{Pruning as Coverage}
\label{sec:coverage}

Most pruning methods answer a local question: \emph{which individual tokens are most important?} We instead ask a set-level question. After retaining a set $S\!\subset\![n]$ of size $m$, is every pruned token $i\!\notin\!S$ adequately represented by some retained $j\!\in\!S$? We define the representational coverage of $i$ under retrained set $S$ as
\begin{equation}
    \mathrm{cov}(i, S) = \max_{j \in S}\; \mathrm{sim}(\mathbf{v}_i, \mathbf{v}_j),
    \label{eq:cov_single}
\end{equation}
where $\mathrm{sim}(\cdot,\cdot)$ is cosine similarity in the projector's output space. Pruning can then be written as the maximization of weighted total coverage:
\begin{equation}
    f(S) = \sum_{i=1}^{n} w_i \cdot \max_{j \in S}\; \mathrm{sim}(\mathbf{v}_i, \mathbf{v}_j),
    \label{eq:weighted_coverage}
\end{equation}
where $w_i\!\geq\!0$ is a per-token demand weight. We refer to maximizing $f(S)$ subject to $|S|=m$ as Representational Coverage Maximization (RCM). With non-negative utilities and weights, this objective is monotone submodular and admits the standard $(1{-}1/e)$ greedy approximation~\cite{nemhauser1978analysis}, motivating the greedy selector used with the coverage utilities below.

Under this objective, a retained token is valuable when it covers demand that is not yet represented by the current subset, rather than when its own score happens to be large. The selector therefore spends budget where local diversity is high and saves it where redundancy is high, without any spatial supervision.

Two design choices remain: how to define similarity, and how to estimate demand. Rather than use external proxies such as CLIP similarity, later-layer saliency~\cite{chen2024fastv}, or vision-encoder heuristics, \method{} reads both signals from the MLLM that will consume the retained tokens.

\subsection{Anchoring the Pruner in the LLM}
\label{sec:query}

Under the coverage formulation, the representation space defines what it means for one token to stand in for another. \method{} therefore anchors both coverage and query relevance inside the consuming VLM, leading to three design choices.

\paragraph{Similarity in LLM space.}
We compute the similarity in Eq.~\ref{eq:weighted_coverage} in the projector output space, $\mathbf{V}\!=\!g(f_v(X))\!\in\!\mathbb{R}^{n\times d}$, where visual tokens have been mapped to the representation consumed by the LLM. Redundancy is therefore defined after multimodal projection: two patches are close if they provide similar projected evidence to $f_\phi$, even when their vision-encoder features are not nearest neighbors. Cosine similarity in this space captures the substitutability relevant to the downstream decoder.

\paragraph{Calibrating the similarity matrix.}
In practice, raw cosine similarities between projector-output ViT patches often have limited dynamic range. The $\max$ in Eq.~\ref{eq:cov_single} relies on dynamic range: when every token looks broadly similar to every other, coverage gains are uniformly inflated and the greedy algorithm cannot distinguish a redundant token from a novel one. We row-wise mean-center the similarity matrix,
\begin{equation}
    \widetilde{M}_{ij} = M_{ij} - \frac{1}{n}\sum_{k=1}^{n} M_{ik}, \quad M_{ij} = \mathrm{sim}(\mathbf{v}_i, \mathbf{v}_j),
    \label{eq:mean_center}
\end{equation}
so that $\widetilde{M}_{ij}$ measures how much more similar tokens $i$ and $j$ are than the average pair from the same image. We then form calibrated coverage utilities $A_{ij}=\max(0,\widetilde{M}_{ij})$, which treat below-average matches as providing no coverage. Clipping keeps the deployed objective in the same non-negative best-representative form as Eq.~\ref{eq:weighted_coverage}, while giving the selection step more discriminative utility values when raw projector-space similarities often concentrate in a narrow band.

\paragraph{Relevance from the LLM's first attention.}
The demand weight $w_i$ should reflect which visual tokens the current instruction asks the model to consult. We estimate this demand with a lightweight attention probe using only the first decoder layer. After applying each model's chat template and processor, we use the text positions corresponding to the user instruction as queries and the visual-token positions as keys. With the layer-0 input LayerNorm and its per-head $W_Q$ and $W_K$, we compute the scaled dot-product attention $\alpha^{(0,h)}_{t,i}$ that instruction position $t$ places on visual token $i$ in head $h$, and average over the instruction positions $\mathcal{T}_q$ and the $H$ heads:
\begin{equation}
    r_i = \frac{1}{|\mathcal{T}_q|H}\sum_{t\in\mathcal{T}_q}\sum_{h=1}^{H} \alpha^{(0,h)}_{t,i}.
    \label{eq:attention_relevance}
\end{equation}
This is not an extracted attention map from a full decoder pass. It is a one-layer relevance probe computed before pruning, after each MLLM has placed visual and text tokens into its native input layout.

We convert $\mathbf{r}$ into non-negative weights through min-max normalisation, $\tilde{r}_i=(r_i\!-\!r_{\min})/(r_{\max}\!-\!r_{\min})\in[0,1]$, followed by an exponential temperature, $w_i = e^{\tau\tilde{r}_i}$. Setting $\tau\!=\!0$ recovers uniform coverage; larger $\tau$ steers more representational budget toward query-relevant regions.

\definecolor{scoringbg}{RGB}{253,234,205}   
\definecolor{mergebg}{RGB}{222,220,250}     
\definecolor{subsetbg}{RGB}{210, 218, 255}    %
\definecolor{sectbg}{RGB}{240,240,240}      
\definecolor{fullmodel}{RGB}{240,240,240}
\definecolor{ratiocolor}{RGB}{0,140,0}    

\begin{table*}[t!]
\caption{\textbf{Performance comparison of LLaVA-1.5-7B under different pruning ratios.} Best in \textbf{bold}.}
\label{tab:main_llava15}
\centering
\footnotesize
\setlength{\tabcolsep}{3pt}
\renewcommand{\arraystretch}{0.92}
\resizebox{0.93\textwidth}{!}{%
\begin{tabular}{l@{\hskip 6pt}cccccccc@{\hskip 6pt}cc}
\toprule
Method & VQA$^{\text{v2}}$ & GQA & SQA$^{\text{I}}$ & VQA$^{\text{T}}$ & POPE & MME
       & MMB$^{\text{EN}}$ & MMB$^{\text{CN}}$ & Acc. & Rel. \\
\midrule
\rowcolor{sectbg}
\multicolumn{11}{c}{\textit{Upper Bound --- All 576 Tokens (\textcolor{ratiocolor}{100\%})}} \\
\rowcolor{fullmodel}
LLaVA-1.5-7B
  & 78.5 & 61.9 & 69.5 & 58.2 & 85.9 & 1506.5 & 64.7 & 58.1 & 69.0 & 100.0\% \\
\midrule
\rowcolor{sectbg}
\multicolumn{11}{c}{\textit{Retain 128 Tokens (\textcolor{ratiocolor}{$\downarrow$77.8\%})}} \\
ToMe~{\scriptsize(ICLR23)}     & 63.0 & 52.4 & 59.6 & 49.1 & 62.8 & 1086.6 & 53.3 & 47.4 & 55.2 & 80.0\% \\
FastV~{\scriptsize(ECCV24)}    & 71.0 & 54.0 & 69.2 & 56.4 & 68.2 & 1368.9 & \textbf{63.0} & 55.9 & 63.3 & 91.7\% \\
PDrop~{\scriptsize(CVPR25)}    & 74.3 & 57.1 & 70.1 & 56.7 & 77.5 & 1444.1 & 62.3 & 55.3 & 65.7 & 95.2\% \\
TRIM~{\scriptsize(COLING25)}   & 75.4 & 58.4 & 68.6 & 52.2 & 85.3 & 1413.4 & \textbf{63.0} & 52.3 & 65.7 & 95.2\% \\
DART~{\scriptsize(EMNLP25)}    & 74.7 & 57.9 & 69.1 & 56.3 & 80.4 & 1408.7 & 60.7 & \textbf{57.3} & 65.9 & 95.4\% \\
PruMerge~{\scriptsize(ICCV25)} & 75.0 & 58.2 & 69.1 & 54.0 & 83.1 & 1408.1 & 61.8 & 55.8 & 65.9 & 95.5\% \\
SparseVLM~{\scriptsize(ICML25)}& 75.1 & 57.3 & 69.0 & 56.3 & 83.1 & 1399.3 & 62.6 & 56.9 & 66.3 & 96.0\% \\
VisionZip~{\scriptsize(CVPR25)}& 75.6 & 57.6 & 68.7 & 56.9 & 83.3 & 1436.9 & 62.1 & 57.0 & 66.6 & 96.5\% \\
DivPrune~{\scriptsize(CVPR25)} & 76.0 & \textbf{59.4} & 68.6 & 55.9 & 87.0 & 1405.1 & 61.5 & 54.8 & 66.7 & 96.6\% \\
MMTok~{\scriptsize(ICLR26)}    & 76.3 & 59.3 & 68.8 & \textbf{57.0} & 86.2 & 1435.2 & 62.3 & 55.4 & 67.1 & 97.3\% \\
\rowcolor{subsetbg}
\textbf{\method{} (Ours)} & \textbf{76.7} & 59.1 & \textbf{70.4} & 56.9 & \textbf{87.5} & \textbf{1465.0} & 62.9 & 56.5 & \textbf{67.9} & \textbf{98.4\%} \\
\midrule
\rowcolor{sectbg}
\multicolumn{11}{c}{\textit{Retain 64 Tokens (\textcolor{ratiocolor}{$\downarrow$88.9\%})}} \\
ToMe~{\scriptsize(ICLR23)}     & 57.1 & 48.6 & 50.0 & 45.3 & 52.5 & 920.7 & 43.7 & 38.5 & 47.7 & 69.1\% \\
PDrop~{\scriptsize(CVPR25)}    & 56.3 & 46.1 & 68.8 & 49.2 & 40.8 & 982.2 & 48.0 & 36.6 & 49.4 & 71.5\% \\
FastV~{\scriptsize(ECCV24)}    & 55.9 & 46.0 & \textbf{70.1} & 51.6 & 35.5 & 973.5 & 50.1 & 42.1 & 50.0 & 72.4\% \\
SparseVLM~{\scriptsize(ICML25)}& 66.9 & 52.0 & 69.2 & 52.1 & 69.7 & 1190.4 & 58.3 & 49.6 & 59.7 & 86.5\% \\
PruMerge~{\scriptsize(ICCV25)} & 71.3 & 55.4 & 69.5 & 52.0 & 75.7 & 1316.8 & 59.6 & 52.1 & 62.7 & 90.8\% \\
DART~{\scriptsize(EMNLP25)}    & 71.3 & 54.7 & 69.3 & 54.7 & 73.8 & 1365.1 & 59.5 & 54.0 & 63.2 & 91.6\% \\
TRIM~{\scriptsize(COLING25)}   & 72.4 & 56.1 & 69.0 & 49.7 & 85.9 & 1350.9 & 60.9 & 48.2 & 63.7 & 92.3\% \\
VisionZip~{\scriptsize(CVPR25)}& 72.4 & 55.1 & 69.0 & 55.5 & 77.0 & 1365.2 & 60.1 & \textbf{55.4} & 64.1 & 92.9\% \\
DivPrune~{\scriptsize(CVPR25)} & 74.1 & 57.5 & 68.0 & 54.5 & 85.5 & 1334.7 & 60.1 & 52.3 & 64.8 & 94.0\% \\
MMTok~{\scriptsize(ICLR26)}    & 75.2 & 58.3 & 69.2 & \textbf{56.0} & 85.8 & 1371.5 & \textbf{61.2} & 53.0 & 66.0 & 95.6\% \\
\rowcolor{subsetbg}
\textbf{\method{} (Ours)} & \textbf{75.9} & \textbf{58.4} & 69.3 & 55.6 & \textbf{86.1} & \textbf{1432.4} & 60.8 & 54.4 & \textbf{66.5} & \textbf{96.4\%} \\
\midrule
\rowcolor{sectbg}
\multicolumn{11}{c}{\textit{Retain 32 Tokens (\textcolor{ratiocolor}{$\downarrow$94.4\%})}} \\
PruMerge~{\scriptsize(ICCV25)} & 65.6 & 52.9 & 69.3 & 49.2 & 66.7 & 1236.6 & 55.1 & 45.9 & 58.3 & 84.5\% \\
VisionZip~{\scriptsize(CVPR25)}& 67.1 & 51.8 & 69.1 & 53.1 & 69.4 & 1251.2 & 57.0 & \textbf{50.3} & 60.0 & 87.0\% \\
DART~{\scriptsize(EMNLP25)}    & 67.1 & 52.9 & 69.3 & 52.2 & 69.1 & 1273.3 & 58.5 & 50.0 & 60.3 & 87.4\% \\
TRIM~{\scriptsize(COLING25)}   & 68.6 & 54.5 & 68.1 & 47.6 & \textbf{84.9} & 1251.8 & 57.7 & 40.1 & 60.5 & 87.7\% \\
DivPrune~{\scriptsize(CVPR25)} & 71.2 & 54.9 & 68.6 & 52.9 & 81.5 & 1284.9 & 57.6 & 49.1 & 62.5 & 90.6\% \\
\rowcolor{subsetbg}
\textbf{\method{} (Ours)} & \textbf{72.3} & \textbf{56.0} & \textbf{69.7} & \textbf{54.3} & 84.7 & \textbf{1407.1} & \textbf{59.3} & \textbf{50.3} & \textbf{64.6} & \textbf{93.7\%} \\
\bottomrule
\end{tabular}
}
\renewcommand{\arraystretch}{1.0}
\end{table*}

\subsection{Demand-Weighted Coverage Selection}
\label{sec:algorithm}
Given the coverage utilities $A_{ij}$ and demand weights $w_i$, \method{} selects the retained set by maximizing
\begin{equation}
    \tilde{f}(S) = \sum_{i=1}^{n} w_i \cdot \max_{j \in S} A_{ij},
    \label{eq:clipped_obj}
\end{equation}
where $A_{ij}=\max(0,\widetilde{M}_{ij})$ measures whether retained token $j$ represents token $i$ better than the average token in the same image. The term $\max_{j\in S}A_{ij}$ records the best representative that token $i$ currently has in the retained set, while $w_i$ assigns more demand to visual tokens that the instruction asks the model to consult.

We optimize Eq.~\ref{eq:clipped_obj}. Let $c_i$ denote the current coverage of token $i$, initialized as $0$ and updated as $c_i=\max_{j\in S}A_{ij}$. Adding candidate token $k$ improves token $i$ only by the uncovered amount $\max(0,A_{ik}-c_i)$, so each step selects

\begin{equation}
    k^* = \arg\max_{k \notin S}\; \sum_{i=1}^{n} w_i \cdot \max\!\big(0,\; A_{ik} - c_i\big),
    \label{eq:coverage_step}
\end{equation}
then updates $c_i\!\leftarrow\!\max(c_i,A_{ik^*})$ for every $i$ and repeats until $|S|=m$. This update makes each choice depend on what remains uncovered rather than on a static token score: once a region is represented, another similar token contributes little, while a token covering missed evidence remains valuable. After selection, \method{} keeps the selected original projected visual tokens, drops the rest, and feeds the retained sequence to the unmodified decoder. Appendix~\ref{app:algorithm_pipeline} gives the pipeline.

\section{Experiments}
\label{sec:experiments}

\subsection{Experimental Setup}

\paragraph{Models and benchmarks.}
We evaluate \method{} primarily on LLaVA-1.5-7B~\cite{liu2024improved}, and additionally on LLaVA-NeXT-7B~\cite{liu2024llavanext} and Qwen2.5-VL-7B~\cite{bai2025qwen25vl}. The benchmarks span general VQA, science and diagram reasoning, OCR/text-rich QA, hallucination, chart understanding, and multilingual or culturally grounded settings. Dataset details are in Appendix~\ref{sec:appendix_benchmarks}; metric and latency details are in Appendix~\ref{sec:appendix_experimentsetup}.

\paragraph{Baselines.}
We compare with token scoring methods (FastV~\cite{chen2024fastv}, PyramidDrop~\cite{xing2024pyramiddrop}, SparseVLM~\cite{zhang2024sparsevlm}, VisionZip~\cite{yang2024visionzip}, DART~\cite{wen2025dart}), token merging methods (ToMe~\cite{bolya2023tome}, PruMerge~\cite{shang2025prumerge}, TRIM~\cite{song2024trim}), and set-level methods (DivPrune~\cite{alvar2025divprune}, MMTok~\cite{dong2026mmtok}). We evaluate LLaVA-1.5-7B at three pruning ratios, plus LLaVA-NeXT-7B and Qwen2.5-VL-7B at matching budgets. All comparisons use matched backbones, resolutions, protocols, and budgets; method-specific parameters follow public/default configs.

\paragraph{Default configuration.}
\method{} uses mean-centered similarity, first-layer attention relevance with $\tau\!=\!0.3$, and sum aggregation unless otherwise specified.

\begin{table*}[t]
\caption{\textbf{Performance comparison of LLaVA-NeXT-7B under different pruning ratios.} Best in \textbf{bold}.}
\label{tab:main_llavanext}
\centering
\footnotesize
\setlength{\tabcolsep}{3pt}
\renewcommand{\arraystretch}{0.92}
\resizebox{0.95\textwidth}{!}{%
\begin{tabular}{l@{\hskip 6pt}cccccccc@{\hskip 6pt}cc}
\toprule
Method & VQA$^{\text{v2}}$ & GQA & SQA$^{\text{I}}$ & VQA$^{\text{T}}$ & POPE & MME
       & MMB$^{\text{EN}}$ & MMB$^{\text{CN}}$ & Acc. & Rel. \\
\midrule
\rowcolor{sectbg}
\multicolumn{11}{c}{\textit{Upper Bound --- All 2{,}880 Tokens (100\%)}} \\
\rowcolor{fullmodel}
LLaVA-NeXT-7B
  & 81.3 & 62.5 & 67.5 & 60.3 & 86.8 & 1511.8 & 65.8 & 57.3 & 69.6 & 100.0\% \\
\midrule
\rowcolor{sectbg}
\multicolumn{11}{c}{\textit{Retain 640 Tokens (\textcolor{ratiocolor}{$\downarrow$77.8\%})}} \\
FastV~{\scriptsize(ECCV24)}    & 77.0 & 58.9 & 67.4 & 58.1 & 79.5 & 1412.6 & 63.1 & 53.5 & 66.0 & 94.9\% \\
PDrop~{\scriptsize(CVPR25)}    & 79.1 & 60.0 & 66.7 & 57.8 & 83.8 & 1475.9 & 64.1 & 55.2 & 67.6 & 97.0\% \\
PruMerge~{\scriptsize(ICCV25)} & 78.2 & 60.8 & 67.8 & 54.9 & 85.3 & 1480.2 & 64.6 & 57.3 & 67.9 & 97.4\% \\
TRIM~{\scriptsize(COLING25)}   & 78.3 & 62.1 & 66.9 & 54.8 & 86.9 & 1471.8 & \textbf{66.8} & 55.8 & 68.1 & 97.8\% \\
DART~{\scriptsize(EMNLP25)}    & 78.3 & 61.3 & \textbf{68.2} & 59.5 & 85.0 & 1450.2 & 64.9 & 57.1 & 68.4 & 98.3\% \\
DivPrune~{\scriptsize(CVPR25)} & 79.3 & 61.9 & 67.8 & 57.0 & 86.9 & 1469.7 & 65.8 & 57.3 & 68.7 & 98.6\% \\
SparseVLM~{\scriptsize(ICML25)}& 79.2 & 61.2 & 67.6 & 59.7 & 85.3 & 1456.8 & 65.9 & \textbf{58.6} & 68.8 & 98.9\% \\
VisionZip~{\scriptsize(CVPR25)}& 79.1 & 61.2 & 68.1 & \textbf{59.9} & 86.0 & \textbf{1493.4} & 65.8 & 58.1 & 69.1 & 99.3\% \\
\rowcolor{subsetbg}
\textbf{\method{} (Ours)} & \textbf{80.1} & \textbf{62.7} & 68.0 & 58.7 & \textbf{87.5} & 1485.2 & 66.5 & 57.8 & \textbf{69.4} & \textbf{99.7\%} \\
\midrule
\rowcolor{sectbg}
\multicolumn{11}{c}{\textit{Retain 320 Tokens (\textcolor{ratiocolor}{$\downarrow$88.9\%})}} \\
FastV~{\scriptsize(ECCV24)}    & 61.5 & 49.8 & 66.6 & 52.2 & 49.5 & 1099.0 & 53.4 & 42.5 & 53.8 & 78.2\% \\
PDrop~{\scriptsize(CVPR25)}    & 66.8 & 50.4 & 66.7 & 49.0 & 60.8 & 1171.5 & 55.5 & 44.7 & 56.6 & 81.6\% \\
TRIM~{\scriptsize(COLING25)}   & 74.9 & 59.9 & 66.2 & 50.2 & 86.5 & 1443.8 & 63.5 & 51.0 & 65.5 & 93.7\% \\
SparseVLM~{\scriptsize(ICML25)}& 74.6 & 57.9 & 67.2 & 56.5 & 76.9 & 1386.1 & 63.1 & \textbf{56.7} & 65.3 & 94.1\% \\
PruMerge~{\scriptsize(ICCV25)} & 75.3 & 58.8 & \textbf{68.1} & 54.0 & 79.5 & 1444.3 & 63.0 & 55.6 & 65.8 & 94.6\% \\
DART~{\scriptsize(EMNLP25)}    & 75.7 & 59.5 & 67.5 & 57.6 & 81.0 & 1419.5 & 64.2 & 55.7 & 66.5 & 95.7\% \\
VisionZip~{\scriptsize(CVPR25)}& 76.2 & 58.9 & 67.5 & \textbf{58.8} & 82.3 & 1397.1 & 63.3 & 55.6 & 66.6 & 95.7\% \\
DivPrune~{\scriptsize(CVPR25)} & 77.2 & 61.1 & 67.7 & 56.2 & 84.7 & 1423.3 & 63.9 & 55.7 & 67.2 & 96.5\% \\
\rowcolor{subsetbg}
\textbf{\method{} (Ours)} & \textbf{78.6} & \textbf{61.8} & 68.0 & 57.8 & \textbf{87.3} & \textbf{1460.8} & \textbf{65.6} & 56.0 & \textbf{68.5} & \textbf{98.4\%} \\
\midrule
\rowcolor{sectbg}
\multicolumn{11}{c}{\textit{Retain 160 Tokens (\textcolor{ratiocolor}{$\downarrow$94.4\%})}} \\
PruMerge~{\scriptsize(ICCV25)} & 70.5 & 56.2 & 66.9 & 50.3 & 71.1 & 1289.6 & 58.0 & 48.9 & 60.8 & 87.5\% \\
TRIM~{\scriptsize(COLING25)}   & 71.0 & 57.4 & 65.5 & 45.8 & 84.8 & 1275.8 & 61.6 & 45.2 & 61.9 & 88.3\% \\
VisionZip~{\scriptsize(CVPR25)}& 71.4 & 55.2 & \textbf{67.9} & 55.0 & 74.9 & 1327.8 & 58.6 & 50.4 & 62.5 & 89.9\% \\
DART~{\scriptsize(EMNLP25)}    & 72.5 & 56.8 & 67.8 & 54.9 & 75.3 & 1325.4 & 62.0 & 53.6 & 63.6 & 91.7\% \\
DivPrune~{\scriptsize(CVPR25)} & 75.0 & 59.3 & 67.1 & 54.1 & 80.0 & 1356.6 & 62.9 & 53.7 & 65.0 & 93.4\% \\
\rowcolor{subsetbg}
\textbf{\method{} (Ours)} & \textbf{76.9} & \textbf{61.0} & 67.8 & \textbf{55.8} & \textbf{86.9} & \textbf{1432.5} & \textbf{64.3} & \textbf{54.0} & \textbf{67.3} & \textbf{96.6\%} \\
\bottomrule
\end{tabular}
}
\renewcommand{\arraystretch}{1.0}
\end{table*}

\subsection{Main Results}
We evaluate \method{} under tighter token budgets, denser high-resolution sequences, and an internally compressed visual stream.

\paragraph{Results on LLaVA-1.5-7B.}
Table~\ref{tab:main_llava15} reports results across three compression rates. \method{} achieves the best overall Acc among all compared methods at every budget, retaining 98.4\%, 96.4\%, and 93.7\% of full-model performance at 128, 64, and 32 tokens. The gains are largest under aggressive compression, where scoring methods drop sharply on POPE and merging methods lose more on TextVQA.

\begin{figure*}[t]
    \centering
    \resizebox{0.92\textwidth}{!}{%
        \includegraphics{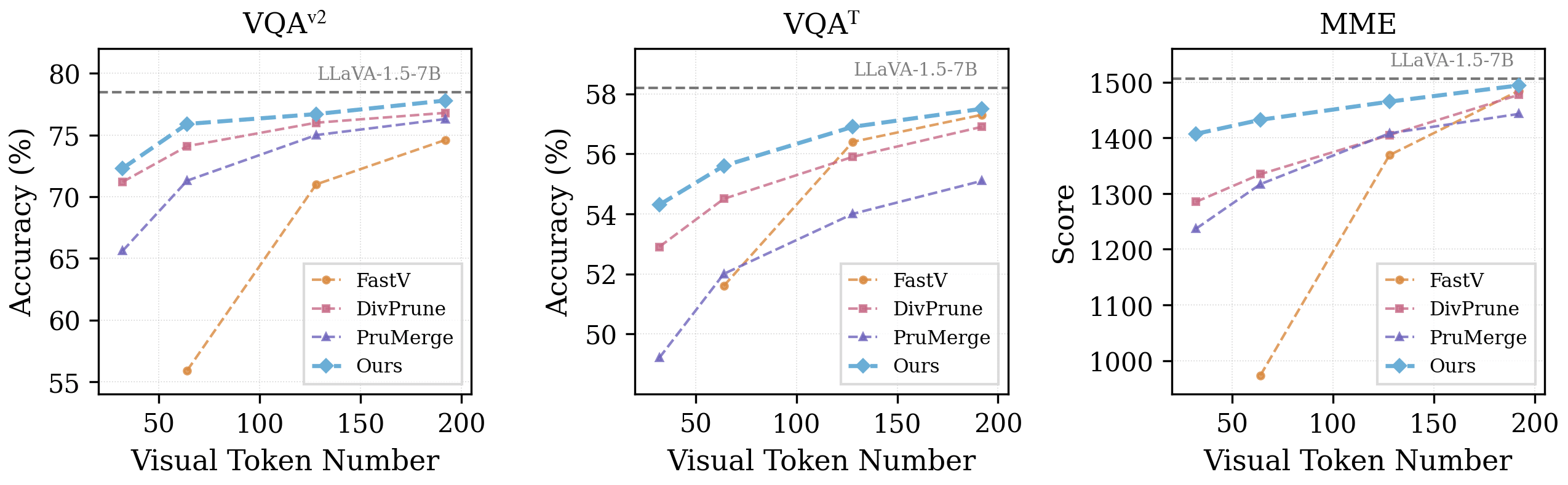}%
    }
    \caption{\textbf{Performance comparison of various methods on LLaVA-1.5-7B at different pruning ratios.}}
    \label{fig:pruning_curves}
    \vspace{-2mm}
\end{figure*}

Figure~\ref{fig:pruning_curves} traces this trend on VQAv2, TextVQA, and MME from 66.7\% to 94.4\% pruning, with \method{} leading at every ratio.

\paragraph{Scaling to high-resolution inputs.}
Table~\ref{tab:main_llavanext} extends the evaluation to LLaVA-NeXT-7B, which produces up to 2,880 visual tokens. \method{} retains 99.7\%, 98.4\%, and 96.6\% of full-model performance at 640, 320, and 160 tokens. Its margin over the strongest baseline grows from 0.3 to 2.3 Acc points as the budget shrinks, suggesting that coverage benefits denser, more redundant visual sequences. The largest gains arise on POPE, where broad scene coverage matters most; even on OCR-heavy TextVQA, \method{} remains the top method at every budget.

\paragraph{Generalization to advanced architectures.}
Table~\ref{tab:main_qwen} evaluates \method{} on Qwen2.5-VL-7B at fixed $1008{\times}1008$ resolution (1,296 visual tokens). Despite Qwen2.5-VL's internal token compression, \method{} achieves the best overall Rel.\ among all Qwen-compatible methods, and its lead grows from 1.3 to 4.1 Rel.\ points across the three pruning levels.

\begin{table*}[t]
\caption{\textbf{Performance comparison on Qwen2.5-VL-7B under different token configurations.} Best in \textbf{bold}.}
\label{tab:main_qwen}
\centering
\footnotesize
\setlength{\tabcolsep}{2pt}
\resizebox{0.98\textwidth}{!}{
\begin{tabular}{l@{\hskip 4pt}cccccccccc@{\hskip 4pt}cc}
\toprule
Method & HallBench & MME & VQA$^{\text{v2}}$ & ChartQA & AI2D & RealWorldQA & CCBench & OCRVQA & SQA$^{\text{I}}$ & POPE & Acc. & Rel. \\
\midrule
\rowcolor{sectbg}
\multicolumn{13}{c}{\textit{Upper Bound --- All 1{,}296 Tokens (\textcolor{ratiocolor}{100\%})}} \\
\rowcolor{fullmodel}
Qwen2.5-VL-7B & 46.8 & 2322.0 & 85.1 & 86.3 & 80.3 & 66.3 & 58.0 & 70.7 & 72.2 & 85.3 & 67.4 & 100.0\% \\
\midrule
\rowcolor{sectbg}
\multicolumn{13}{c}{\textit{Retain 512 Tokens (\textcolor{ratiocolor}{$\downarrow$60.5\%})}} \\
DivPrune~{\scriptsize(CVPR25)} & 43.5 & 2258.0 & 81.8 & 80.3 & 78.7 & 63.8 & 55.1 & 69.0 & 71.0 & 84.4 & 65.0 & 96.3\% \\
FastV~{\scriptsize(ECCV24)}    & 42.4 & \textbf{2318.0} & 84.1 & 82.7 & 78.8 & 64.7 & 55.8 & 67.8 & 71.5 & 84.0 & 65.5 & 97.0\% \\
\rowcolor{subsetbg}
\textbf{\method{} (Ours)} & \textbf{44.5} & 2310.0 & \textbf{84.5} & \textbf{83.5} & \textbf{79.4} & \textbf{65.4} & \textbf{56.6} & \textbf{69.7} & \textbf{71.8} & \textbf{84.6} & \textbf{66.3} & \textbf{98.3\%} \\
\midrule
\rowcolor{sectbg}
\multicolumn{13}{c}{\textit{Retain 256 Tokens (\textcolor{ratiocolor}{$\downarrow$80.2\%})}} \\
DivPrune~{\scriptsize(CVPR25)} & 39.8 & 2162.0 & 75.8 & 69.0 & 76.4 & 60.5 & 53.5 & 65.4 & 70.1 & \textbf{83.5} & 61.6 & 91.3\% \\
FastV~{\scriptsize(ECCV24)}    & 39.0 & 2233.0 & 81.5 & 71.4 & 76.2 & 62.5 & 54.1 & 65.6 & \textbf{70.9} & 81.2 & 62.5 & 92.7\% \\
\rowcolor{subsetbg}
\textbf{\method{} (Ours)} & \textbf{41.5} & \textbf{2240.0} & \textbf{82.4} & \textbf{74.0} & \textbf{77.5} & \textbf{63.2} & \textbf{55.0} & \textbf{67.5} & 70.6 & 83.0 & \textbf{63.7} & \textbf{94.5\%} \\
\midrule
\rowcolor{sectbg}
\multicolumn{13}{c}{\textit{Retain 128 Tokens (\textcolor{ratiocolor}{$\downarrow$90.1\%})}} \\
DivPrune~{\scriptsize(CVPR25)} & 33.4 & 1979.0 & 66.6 & 52.5 & 72.1 & 57.9 & 48.4 & 57.8 & 68.6 & \textbf{81.8} & 55.9 & 82.9\% \\
FastV~{\scriptsize(ECCV24)}    & 33.8 & 2028.0 & 73.8 & 54.3 & 71.4 & 57.8 & 46.7 & 62.7 & 69.4 & 75.8 & 56.6 & 83.9\% \\
\rowcolor{subsetbg}
\textbf{\method{} (Ours)} & \textbf{36.8} & \textbf{2095.0} & \textbf{76.5} & \textbf{60.5} & \textbf{74.0} & \textbf{59.4} & \textbf{50.5} & \textbf{64.8} & \textbf{69.7} & 80.0 & \textbf{59.3} & \textbf{88.0\%} \\
\bottomrule
\end{tabular}
}
\vspace{-2mm}
\end{table*}



\subsection{Ablation Studies}
\label{sec:ablation}
On LLaVA-1.5-7B with 64 retained tokens, we ablate two design choices on TextVQA and POPE (Table~\ref{tab:ablation}): how the relevance signal enters the selection (ranking vs.\ demand weight), and where that signal is read from. Ranking and coverage are not interchangeable: raw-attention ranking reduces POPE substantially below the full-model score, whereas applying the same signal as a demand weight inside coverage recovers full-model-level performance. A relevance score alone identifies tokens aligned with the instruction but cannot ensure that the retained set spans the visual evidence the model needs; used as a demand weight inside coverage, it allocates budget to relevant regions while allowing redundant ones to share representatives.

Within coverage, the relevance source also matters: replacing CLIP with an LLM embedding significantly improves TextVQA, and the first-layer attention probe yields the best result on both benchmarks. The strongest configuration thus combines an LLM-anchored relevance signal with the coverage objective. Further ablations on overhead, similarity calibration, demand temperature, and aggregation variants are reported in Appendix~\ref{sec:appendix_ablation}.

\begin{table}[h!]
\caption{\textbf{Ablation on coverage and relevance signal} (LLaVA-1.5-7B, 64 retained tokens).}
\label{tab:ablation}
\centering
\small
\setlength{\tabcolsep}{6pt}
\resizebox{0.95\columnwidth}{!}{%
\begin{tabular}{lcc}
\toprule
Configuration & TextVQA & POPE \\
\midrule
Full model (576 tokens) & 58.2 & 85.9 \\
\midrule
Top-$m$ by attention rel.\ & 51.9 & 73.0 \\
\midrule
Coverage + CLIP rel.\ & 47.3 & 83.8 \\
Coverage + LLM-embed rel.\ & 55.2 & 85.6 \\
Coverage + Attention rel.\ (\textbf{Ours}) & \textbf{55.6} & \textbf{86.1} \\
\bottomrule
\end{tabular}
}
\end{table}

\begin{table}[h!]
\caption{\textbf{Efficiency comparisons on POPE.} (LLaVA-NeXT-7B at 320 retained tokens).}
\label{tab:efficiency}
\centering
\small
\setlength{\tabcolsep}{4pt}
\resizebox{\columnwidth}{!}{%
\begin{tabular}{lcccc}
\toprule
Methods & Prefill Time $\downarrow$ & Decode Time $\downarrow$ & FLOPs $\downarrow$ & F1 $\uparrow$ \\
\midrule
LLaVA-NeXT-7B       & 246 ms & 52 ms & 41.7 T & 86.8 \\
\midrule
SparseVLM          & 71 ms & 25 ms & 4.5 T & 76.9 \\
DivPrune           & 38 ms & 22 ms & 4.2 T & 84.7 \\
\textbf{\method{}} & \textbf{33 ms} & \textbf{20 ms} & \textbf{3.9 T} & \textbf{87.3} \\
\bottomrule
\end{tabular}%
}
\end{table}

\subsection{Efficiency Analysis}

Beyond accuracy, Table~\ref{tab:efficiency} compares efficiency on POPE with LLaVA-NeXT-7B, using the full 2,880-token model as the upper bound and 320 retained tokens for pruning methods. Prefill and Decode Time are per-token decoder latencies. \method{} reduces prefill latency by 7.5$\times$, decode latency by 2.6$\times$, and FLOPs by 10.7$\times$ while also improving F1 ($86.8 \to 87.3$). It also outperforms SparseVLM and DivPrune on both latency and F1; Appendix~\ref{app:overhead_breakdown} breaks down overhead.

\begin{figure*}[h!]
    \centering
    \resizebox{\textwidth}{!}{%
        \includegraphics{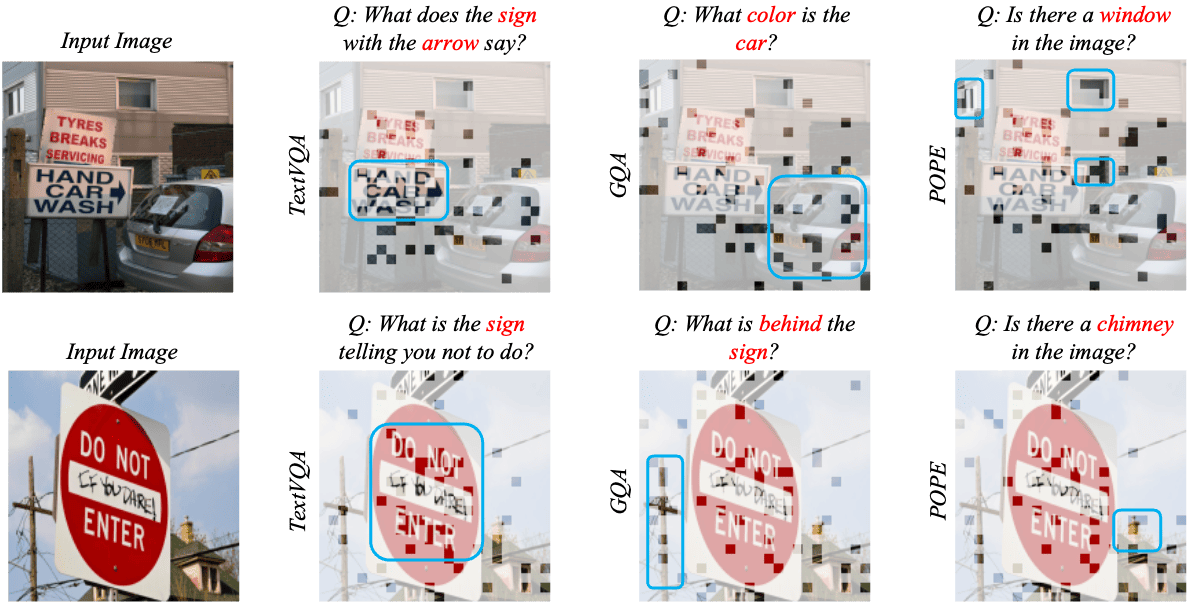}%
    }
    \caption{\textbf{Token selection visualization.} Coverage produces density-adaptive allocation: homogeneous regions receive sparse representation, while informative regions (text, queried objects, semantic labels) receive denser coverage, all without explicit spatial supervision.}
    \label{fig:qualitative}
\end{figure*}

\subsection{Qualitative Analysis}
\label{sec:visualization}

Figure~\ref{fig:qualitative} visualizes \method{}'s token selection on three benchmarks. The selector does not form a single saliency cluster: it concentrates retained tokens on text-bearing regions for TextVQA, on semantically informative objects and their spatial relationships for GQA, and spreads them across the whole scene for POPE. The pattern emerges without explicit spatial supervision: diverse regions receive more representatives while redundant ones share one, with query relevance shaping allocation rather than ordering tokens.

\section{Analysis and Discussion}
\label{sec:discussion}

\paragraph{Ranking and coverage are not interchangeable.}
The ablation shows that first-layer attention has different effects under different objectives. Used as a ranker, it concentrates retained tokens around query-salient patches and leaves much of the image unrepresented; used as a coverage weight, it shifts budget toward relevant regions while preserving representatives for each token. Thus, relevance is most useful as demand inside a set objective, not as an isolated token score~\cite{li2026why}.

\paragraph{Where coverage helps most.}
  \method{} gains most when the retained budget is tight. At moderate budgets, several subsets can still preserve enough visual evidence, so method gaps are smaller. Under aggressive compression, the failure modes of existing methods become more visible: attention-based methods often over-concentrate on a few salient patches, similarity-based methods remain query-blind, and merging methods alter the token representations passed to the decoder. A set-level coverage objective instead selects tokens by their marginal contribution to uncovered evidence, so its advantage is most pronounced under constrained budgets. The qualitative patterns show the same behaviour spatially: \method{} assigns more tokens to diverse evidence and leaves repetitive regions sparse, which is why a shorter prefill does not force an accuracy collapse as long as the retained set stays representative. 

\paragraph{Reading signals from the consuming LLM.}
The token-selection signals in \method{} are all read from the model that will consume the retained tokens. Projector-space similarity measures redundancy after the visual encoder has been translated into the LLM input space, mean-centering makes that similarity useful for best-representative selection, and first-layer attention supplies query demand before pruning. This avoids external proxies and helps a single configuration transfer across benchmarks and architectures. Viewed this way, visual-token pruning is closer to set construction than to ranking, and a relevance signal becomes useful only after the objective fixes which visual evidence must remain represented.

\section{Conclusion}
\label{sec:conclusion}

We presented \method{}, a training-free pruner that formulates visual-token selection as Representational Coverage Maximization. \method{} combines projector-space coverage, mean-centered non-negative utilities, and a first-layer attention probe to select original visual tokens for the consuming LLM, without modifying the model or adding any trained component. Across the evaluated VLMs and compression rates, it achieves the highest average accuracy among the baselines we compare, and its margin widens as the token budget tightens. The same regime where coverage helps most is also where existing rankers and merging methods break down, suggesting that aggressive pruning should choose representatives for the evidence that disappears rather than rank the evidence that remains. Within this view, query relevance contributes more by reshaping demand than by ordering tokens.

\section{Limitations}
\label{sec:limitations}
\method{} requires access to projector visual tokens and early decoder signals, which limits its direct use with closed API-only MLLMs. Coverage holds well for redundant scene regions, repeated textures, and broad contextual cues, but is weaker when the answer depends on small objects, OCR details, precise counting, spatial relations, or other local evidence that may not be replaceable even when projected-token similarity is high. In addition, pairwise similarity is quadratic in the number of visual tokens; while practical for the image resolutions evaluated here, higher-resolution inputs require more careful engineering. This risk is especially relevant in accessibility, safety-sensitive, multilingual, and culturally grounded settings, where fine-grained visual evidence can be consequential.

Unlike single images, video redundancy is both spatial and temporal, and the evidence needed for an answer may be a short-lived action, a transient object, or a change between frames. Applying image-level coverage directly to all video tokens could over-represent persistent background evidence while under-representing temporally sparse events. Similarly, \method{} performs selection once before decoding, which matches many VQA settings but does not capture cases where visual demand shifts during multi-step reasoning, dialogue, or long-form generation. Beyond average benchmark retention, finer-grained analysis across OCR, counting, grounding, spatial reasoning, hallucination, and temporal localization would better characterize when projected-space coverage behaves as true evidence substitution.

\section*{Acknowledgment}
This work is partly supported by the National Science Foundation IIS-2223768 and DRL-2507128.

\bibliography{ref}

\clearpage

\appendix
\section*{Appendix}

\section{Experimental Setup and Benchmark Details}

\subsection{Experiment Setup}
\label{sec:appendix_experimentsetup}
This section summarizes the evaluation benchmarks used in the main experiments. We include only the datasets that appear in our tables and follow the official setting and metric for each benchmark. Latency is measured on a single NVIDIA RTX A6000 GPU with FP16 precision, averaged over 100 forward passes after 10 warm-up iterations. Decoder prefill/decode latencies are reported per token and exclude tokenization, vision encoding, and projection. Acc.\ denotes the benchmark average and Rel.\ the percentage of full-model Acc.\ retained.

\subsection{Benchmark Details}
\label{sec:appendix_benchmarks}
\paragraph{VQAv2.}
VQAv2~\cite{goyal2017making} evaluates open-ended visual question answering on natural images. The dataset was designed to reduce language-prior shortcuts by pairing visually similar images with different answers, making it a broad test of whether a model preserves enough visual evidence after pruning.

\paragraph{GQA.}
GQA~\cite{hudson2019gqa} emphasizes compositional scene understanding. Its questions are grounded in scene-graph annotations and often require reasoning over objects, attributes, and relations, so it is useful for measuring whether token reduction preserves structured visual content rather than only salient foreground regions.

\paragraph{ScienceQA-IMG.}
ScienceQA-IMG~\cite{lu2022learn} contains science questions paired with images or diagrams. We use its image subset to test multimodal reasoning under visual-token budgets, especially cases where the answer depends on small diagrammatic details or localized evidence.

\paragraph{TextVQA.}
TextVQA~\cite{singh2019towards} evaluates question answering over images that contain readable text. Because OCR-relevant patches are often small and spatially scattered, this benchmark is particularly sensitive to whether pruning retains locally distinctive visual tokens.

\paragraph{POPE.}
POPE~\cite{li2023evaluating} probes object hallucination by asking binary questions about whether specific objects are present in an image. It stresses coverage of the whole scene: pruning methods that over-concentrate on a few salient regions can lose contextual evidence needed for object-presence judgments.

\paragraph{MME.}
MME~\cite{fu2026mme} provides a broad instruction-style evaluation of perceptual and reasoning abilities. It reports scores over multiple sub-tasks rather than a single percentage accuracy; when computing average accuracy in our tables, we normalize MME by the corresponding full score used by the evaluation protocol.

\paragraph{MMBench and MMBench-CN.}
MMBench~\cite{liu2024mmbench} evaluates multimodal models through multiple-choice questions spanning perception and reasoning skills. We report both the English and Chinese versions, denoted as MMBench and MMBench-CN, to measure whether pruning behavior transfers across language settings.

\paragraph{ChartQA.}
ChartQA~\cite{masry2022chartqa} focuses on question answering over charts. It requires models to read visual marks and textual labels while performing basic logical or numerical reasoning, making it a useful stress test for high-resolution models such as Qwen2.5-VL.

\paragraph{AI2D.}
AI2D~\cite{kembhavi2016diagram} evaluates understanding of scientific diagrams. Compared with natural-image VQA, it places more weight on schematic structure, labels, arrows, and part-whole relations, providing a complementary test of whether token pruning preserves diagram-level information.

\paragraph{HallBench.}
HallBench~\cite{guan2024hallusionbench} evaluates visual hallucination and illusion failures in large vision-language models. It is useful for testing whether pruning preserves the image evidence needed to avoid visually unsupported answers.

\paragraph{RealWorldQA.}
RealWorldQA~\cite{xai2024grok} evaluates real-world spatial understanding and perception. Its questions stress practical scene interpretation, making it a complementary high-resolution benchmark for Qwen2.5-VL.

\paragraph{CCBench.}
CCBench~\cite{liu2024mmbench} focuses on Chinese cultural contexts and knowledge. We include it in the Qwen2.5-VL evaluation to test whether pruning preserves culturally grounded visual evidence beyond generic scene recognition.

\paragraph{OCRVQA.}
OCRVQA~\cite{mishra2019ocrvqa} evaluates visual question answering over text-rich images such as book covers. Alongside TextVQA and ChartQA, it stresses whether retained tokens preserve fine-grained textual evidence.

\section{\method{} Algorithm Pipeline}
\label{app:algorithm_pipeline}

Algorithm~\ref{alg:coverpruner} summarizes the inference-time pipeline used by \method{}. It follows the components in Section~\ref{sec:method}: projected visual-token coverage, mean-centered non-negative utilities, first-layer attention demand, and demand-weighted coverage selection.

\begin{algorithm}[h!]
\caption{\method{} inference-time pruning pipeline}
\label{alg:coverpruner}
\small
\setlength{\tabcolsep}{3pt}
\renewcommand{\arraystretch}{1.12}
\begin{tabular}{@{}p{0.18\columnwidth}p{0.76\columnwidth}@{}}
\toprule
\textbf{Input} & Image $X_v$, instruction $q$, VLM $(f_v,g,f_\phi)$, budget $m$, temperature $\tau$ \\
\textbf{Output} & Retained visual-token sequence $\mathbf{V}_S$ \\
\midrule
\multicolumn{2}{@{}l}{\textbf{1. Build coverage utilities}} \\
Preprocess & Apply the model processor and chat template to obtain $\mathbf{H}_q$, visual-token positions, and instruction positions $\mathcal{T}_q$. \\
Project & Compute $\mathbf{V}=g(f_v(X_v))=\{\mathbf{v}_i\}_{i=1}^{n}$. \\
Cover & Compute $M_{ij}=\mathrm{sim}(\mathbf{v}_i,\mathbf{v}_j)$ and
$A_{ij}=\max(0,M_{ij}-\frac{1}{n}\sum_{k=1}^{n}M_{ik})$. \\
\midrule
\multicolumn{2}{@{}l}{\textbf{2. Estimate query demand}} \\
Probe & Use first-layer instruction queries in $\mathcal{T}_q$ and visual-token keys to obtain attention scores. \\
Weight & Average over instruction positions and heads to get $r_i$ as in Eq.~\ref{eq:attention_relevance}; set
$w_i=\exp\!\big(\tau(r_i-r_{\min})/(r_{\max}-r_{\min})\big)$. \\
\midrule
\multicolumn{2}{@{}l}{\textbf{3. Select retained tokens}} \\
Initialize & Set $S=\emptyset$ and $c_i=0$ for all visual tokens. \\
Iterate & While $|S|<m$, choose
$k^*=\arg\max_{k\notin S}\sum_i w_i\max(0,A_{ik}-c_i)$; then update
$S\leftarrow S\cup\{k^*\}$ and $c_i\leftarrow\max(c_i,A_{ik^*})$. \\
Decode & Return $\mathbf{V}_S$ in the original visual-token order and feed $[\mathbf{V}_S;\mathbf{H}_q]$ to the unchanged decoder $f_\phi$. \\
\bottomrule
\end{tabular}
\renewcommand{\arraystretch}{1.0}
\end{algorithm}

\section{Additional Experimental Results}
\label{app:additional}

In the main paper, we present experiments on LLaVA-1.5-7B and LLaVA-NeXT-7B. To further
demonstrate the generalizability of our method across model scales, we provide additional results
on LLaVA-1.5-13B and LLaVA-NeXT-13B.

\subsection{\method{} on LLaVA-1.5-13B}
\label{app:llava}
\label{app:llava15_13b}

Table~\ref{tab:app_llava13b} presents the full per-benchmark results for LLaVA-1.5-13B across all three compression rates. \method{} achieves the best overall accuracy among all compared methods under each token budget. With 77.8\% of visual tokens removed, it retains 98.4\% of the full-model performance; under the most aggressive 94.4\% reduction, the Rel.\ gap over the strongest competing method widens to 2.9 points. This trend mirrors the main-result setting and shows that the coverage objective remains useful as model scale increases.

\begin{table*}[h!]
\caption{\textbf{Performance comparison of LLaVA-1.5-13B under different pruning ratios.}}
\label{tab:app_llava13b}
\centering
\footnotesize
\setlength{\tabcolsep}{3pt}
\renewcommand{\arraystretch}{0.96}
\resizebox{0.82\textwidth}{!}{%
\begin{tabular}{l@{\hskip 6pt}cccccccc@{\hskip 6pt}cc}
\toprule
Method & VQA$^{\text{v2}}$ & GQA & SQA$^{\text{I}}$ & VQA$^{\text{T}}$ & POPE & MME
       & MMB$^{\text{EN}}$ & MMB$^{\text{CN}}$ & Acc. & Rel. \\
\midrule
\rowcolor{sectbg}
\multicolumn{11}{c}{\textit{Upper Bound --- All 576 Tokens (\textcolor{ratiocolor}{100\%})}} \\
\rowcolor{fullmodel}
LLaVA-1.5-13B
  & 80.0 & 63.3 & 72.8 & 61.2 & 86.0 & 1531.2 & 68.5 & 63.5 & 71.5 & 100.0\% \\
\midrule
\rowcolor{sectbg}
\multicolumn{11}{c}{\textit{Retain 128 Tokens (\textcolor{ratiocolor}{$\downarrow$77.8\%})}} \\
FastV~{\scriptsize(ECCV24)}    & 75.3 & 58.3 & 74.2 & 58.6 & 75.5 & 1460.6 & 66.1 & 62.3 & 67.9 & 95.0\% \\
PDrop~{\scriptsize(CVPR25)}    & 78.2 & \textbf{61.0} & 73.3 & \textbf{60.2} & 83.6 & 1489.5 & 67.5 & \textbf{62.8} & 70.1 & 98.1\% \\
SparseVLM~{\scriptsize(ICML25)}& 77.6 & 59.6 & 74.3 & 59.3 & 85.0 & 1487.9 & \textbf{68.4} & 62.6 & 70.1 & 98.1\% \\
VisionZip~{\scriptsize(CVPR25)}& 76.8 & 57.9 & 73.8 & 58.9 & 82.7 & 1449.2 & 67.4 & 62.5 & 69.1 & 96.6\% \\
DART~{\scriptsize(EMNLP25)}    & 75.7 & 57.7 & 74.2 & 58.7 & 80.4 & 1395.0 & 65.4 & 62.2 & 68.0 & 95.1\% \\
PruMerge~{\scriptsize(ICCV25)} & 76.2 & 58.3 & 73.3 & 56.1 & 82.7 & 1445.9 & 66.3 & 61.2 & 68.3 & 95.5\% \\
TRIM~{\scriptsize(COLING25)}   & 76.4 & 59.4 & 72.4 & 55.0 & 86.8 & 1426.9 & 67.1 & 58.4 & 68.4 & 95.6\% \\
DivPrune~{\scriptsize(CVPR25)} & 77.1 & 59.2 & 72.8 & 58.0 & 86.8 & 1457.7 & 66.3 & 60.7 & 69.2 & 96.8\% \\
MMTok~{\scriptsize(ICLR26)}    & 77.5 & 58.9 & 73.3 & 59.2 & 86.1 & 1452.3 & 67.1 & 60.5 & 69.4 & 97.1\% \\
\rowcolor{subsetbg}
\textbf{\method{} (Ours)} & \textbf{78.7} & 59.7 & \textbf{74.5} & 58.9 & \textbf{87.3} & \textbf{1490.5} & 67.5 & 61.5 & \textbf{70.3} & \textbf{98.4\%} \\
\midrule
\rowcolor{sectbg}
\multicolumn{11}{c}{\textit{Retain 64 Tokens (\textcolor{ratiocolor}{$\downarrow$88.9\%})}} \\
FastV~{\scriptsize(ECCV24)}    & 65.3 & 51.9 & 73.1 & 53.4 & 56.9 & 1246.4 & 59.2 & 55.1 & 59.7 & 83.5\% \\
PDrop~{\scriptsize(CVPR25)}    & 70.8 & 54.1 & 73.1 & 55.3 & 66.1 & 1247.0 & 63.1 & 56.6 & 62.7 & 87.8\% \\
SparseVLM~{\scriptsize(ICML25)}& 73.2 & 55.9 & 73.0 & 57.1 & 77.9 & 1374.3 & 65.2 & 60.3 & 66.4 & 93.0\% \\
VisionZip~{\scriptsize(CVPR25)}& 73.7 & 56.2 & \textbf{74.2} & 57.4 & 75.7 & 1379.6 & 64.9 & \textbf{61.3} & 66.5 & 93.2\% \\
DART~{\scriptsize(EMNLP25)}    & 72.4 & 55.7 & 73.8 & 57.3 & 72.8 & 1380.0 & 64.7 & 60.6 & 65.8 & 92.1\% \\
PruMerge~{\scriptsize(ICCV25)} & 72.6 & 56.3 & 73.5 & 54.4 & 75.7 & 1338.2 & 65.0 & 59.3 & 65.5 & 91.7\% \\
TRIM~{\scriptsize(COLING25)}   & 73.2 & 57.9 & 72.0 & 52.0 & 86.5 & 1406.2 & 65.0 & 52.7 & 66.2 & 92.7\% \\
DivPrune~{\scriptsize(CVPR25)} & 75.2 & 57.9 & 71.7 & 57.4 & 84.5 & 1454.2 & 64.1 & 59.8 & 67.9 & 95.1\% \\
MMTok~{\scriptsize(ICLR26)}    & 76.3 & 58.4 & 72.9 & \textbf{58.4} & 84.3 & 1431.2 & \textbf{65.7} & 57.6 & 68.1 & 95.3\% \\
\rowcolor{subsetbg}
\textbf{\method{} (Ours)} & \textbf{76.7} & \textbf{59.4} & 72.5 & 58.0 & \textbf{87.1} & \textbf{1466.8} & 65.5 & 58.8 & \textbf{68.9} & \textbf{96.5\%} \\
\midrule
\rowcolor{sectbg}
\multicolumn{11}{c}{\textit{Retain 32 Tokens (\textcolor{ratiocolor}{$\downarrow$94.4\%})}} \\
VisionZip~{\scriptsize(CVPR25)}& 68.4 & 52.7 & 72.9 & 55.2 & 66.8 & 1257.7 & 61.2 & 55.8 & 62.0 & 86.8\% \\
DART~{\scriptsize(EMNLP25)}    & 68.1 & 53.9 & \textbf{73.2} & 55.1 & 66.9 & 1282.8 & 61.9 & 56.2 & 62.4 & 87.4\% \\
PruMerge~{\scriptsize(ICCV25)} & 66.8 & 54.1 & 71.7 & 52.4 & 67.4 & 1269.1 & 61.1 & 53.5 & 61.3 & 85.9\% \\
TRIM~{\scriptsize(COLING25)}   & 69.8 & 55.6 & 70.4 & 49.6 & 85.8 & 1284.7 & 63.1 & 45.4 & 63.0 & 88.2\% \\
DivPrune~{\scriptsize(CVPR25)} & 72.0 & 56.2 & 70.9 & 54.6 & 79.3 & 1405.2 & 61.7 & \textbf{57.2} & 65.3 & 91.4\% \\
\rowcolor{subsetbg}
\textbf{\method{} (Ours)} & \textbf{75.2} & \textbf{58.5} & 71.9 & \textbf{55.3} & \textbf{86.6} & \textbf{1421.0} & \textbf{63.7} & 56.6 & \textbf{67.4} & \textbf{94.3\%} \\
\bottomrule
\end{tabular}%
}
\end{table*}

\subsection{\method{} on LLaVA-NeXT-13B}
\label{app:llavanext_13b}

Table~\ref{tab:app_llavanext13b} reports the corresponding results for LLaVA-NeXT-13B, whose higher-resolution visual token sequence creates a denser pruning target. \method{} again achieves the best overall accuracy among all compared methods at every budget, retaining 99.9\%, 98.3\%, and 96.8\% of full-model performance as the retained-token count decreases. The margin over the strongest baseline increases under heavier compression, suggesting that coverage-based selection becomes more valuable when each retained token must represent a larger portion of the original visual sequence.

\begin{table*}[h!]
\caption{\textbf{Performance comparison of LLaVA-NeXT-13B under different pruning ratios.}}
\label{tab:app_llavanext13b}
\centering
\footnotesize
\setlength{\tabcolsep}{3pt}
\renewcommand{\arraystretch}{0.96}
\resizebox{0.82\textwidth}{!}{%
\begin{tabular}{l@{\hskip 6pt}cccccccc@{\hskip 6pt}cc}
\toprule
Method & VQA$^{\text{v2}}$ & GQA & SQA$^{\text{I}}$ & VQA$^{\text{T}}$ & POPE & MME
       & MMB$^{\text{EN}}$ & MMB$^{\text{CN}}$ & Acc. & Rel. \\
\midrule
\rowcolor{sectbg}
\multicolumn{11}{c}{\textit{Upper Bound --- All 2{,}880 Tokens (100\%)}} \\
\rowcolor{fullmodel}
LLaVA-NeXT-13B
  & 82.3 & 64.4 & 73.1 & 63.2 & 85.3 & 1539.5 & 68.5 & 61.2 & 71.9 & 100.0\% \\
\midrule
\rowcolor{sectbg}
\multicolumn{11}{c}{\textit{Retain 640 Tokens (\textcolor{ratiocolor}{$\downarrow$77.8\%})}} \\
FastV~{\scriptsize(ECCV24)}    & 79.4 & 60.9 & 71.7 & 60.7 & 80.2 & 1516.7 & 65.5 & 59.9 & 69.3 & 96.4\% \\
PDrop~{\scriptsize(CVPR25)}    & 81.1 & 62.8 & 71.7 & 62.1 & 84.4 & 1559.1 & 66.6 & 60.8 & 70.9 & 98.7\% \\
SparseVLM~{\scriptsize(ICML25)}& 79.9 & 62.7 & \textbf{72.5} & \textbf{62.8} & 85.6 & \textbf{1562.7} & 68.8 & \textbf{64.0} & 71.7 & 99.8\% \\
DART~{\scriptsize(EMNLP25)}    & 79.3 & 62.7 & 71.0 & 61.3 & 85.2 & 1542.4 & 67.6 & 61.9 & 70.8 & 98.5\% \\
VisionZip~{\scriptsize(CVPR25)}& 79.7 & 62.9 & 70.8 & 62.1 & 85.8 & 1549.2 & 68.1 & 62.6 & 71.2 & 99.0\% \\
PruMerge~{\scriptsize(ICCV25)} & 78.7 & 62.8 & 70.6 & 56.2 & 83.7 & 1497.3 & 67.4 & 61.9 & 69.5 & 96.7\% \\
TRIM~{\scriptsize(COLING25)}   & 79.4 & 63.1 & 71.2 & 57.6 & 87.3 & 1554.6 & \textbf{68.9} & 61.2 & 70.8 & 98.5\% \\
DivPrune~{\scriptsize(CVPR25)} & 80.4 & 63.5 & 72.2 & 59.2 & 86.5 & 1526.1 & 67.5 & 62.9 & 71.1 & 98.9\% \\
\rowcolor{subsetbg}
\textbf{\method{} (Ours)} & \textbf{81.9} & \textbf{64.0} & 71.8 & 61.0 & \textbf{87.5} & 1545.6 & \textbf{68.9} & 62.1 & \textbf{71.8} & \textbf{99.9\%} \\
\midrule
\rowcolor{sectbg}
\multicolumn{11}{c}{\textit{Retain 320 Tokens (\textcolor{ratiocolor}{$\downarrow$88.9\%})}} \\
FastV~{\scriptsize(ECCV24)}    & 69.8 & 54.6 & 70.5 & 55.4 & 63.6 & 1279.0 & 59.8 & 54.4 & 61.5 & 85.6\% \\
PDrop~{\scriptsize(CVPR25)}    & 75.4 & 57.7 & 72.1 & 56.2 & 74.6 & 1386.3 & 62.8 & 55.3 & 65.4 & 91.0\% \\
SparseVLM~{\scriptsize(ICML25)}& 76.7 & 60.9 & 70.9 & 60.0 & 81.5 & 1491.6 & \textbf{68.0} & \textbf{63.5} & 69.5 & 96.7\% \\
DART~{\scriptsize(EMNLP25)}    & 76.4 & 60.9 & 69.8 & 59.7 & 81.1 & 1457.4 & 65.9 & 61.9 & 68.6 & 95.4\% \\
VisionZip~{\scriptsize(CVPR25)}& 76.8 & 60.7 & 70.2 & \textbf{60.7} & 82.3 & 1487.3 & 66.5 & 62.3 & 69.2 & 96.3\% \\
PruMerge~{\scriptsize(ICCV25)} & 75.9 & 61.1 & 70.7 & 55.9 & 79.1 & 1426.5 & 66.6 & 60.6 & 67.7 & 94.1\% \\
TRIM~{\scriptsize(COLING25)}   & 75.9 & 61.3 & 69.9 & 52.8 & 87.2 & 1476.6 & 67.3 & 57.4 & 68.2 & 94.9\% \\
DivPrune~{\scriptsize(CVPR25)} & 78.1 & 61.8 & \textbf{72.3} & 57.6 & 85.2 & 1473.0 & 65.9 & 61.9 & 69.6 & 96.8\% \\
\rowcolor{subsetbg}
\textbf{\method{} (Ours)} & \textbf{79.6} & \textbf{63.1} & 71.6 & 58.7 & \textbf{87.6} & \textbf{1498.5} & \textbf{68.0} & 61.8 & \textbf{70.7} & \textbf{98.3\%} \\
\midrule
\rowcolor{sectbg}
\multicolumn{11}{c}{\textit{Retain 160 Tokens (\textcolor{ratiocolor}{$\downarrow$94.4\%})}} \\
DART~{\scriptsize(EMNLP25)}    & 72.8 & 58.7 & 70.1 & 57.2 & 75.7 & 1389.3 & 64.6 & 60.8 & 66.2 & 92.1\% \\
VisionZip~{\scriptsize(CVPR25)}& 72.4 & 57.8 & 69.7 & \textbf{58.6} & 76.8 & 1393.9 & 64.8 & 60.0 & 66.2 & 92.1\% \\
PruMerge~{\scriptsize(ICCV25)} & 71.6 & 57.9 & 70.1 & 52.8 & 72.1 & 1345.9 & 63.2 & 57.1 & 64.0 & 89.1\% \\
TRIM~{\scriptsize(COLING25)}   & 72.1 & 58.9 & 69.1 & 49.2 & 87.0 & 1392.3 & 65.7 & 51.6 & 65.4 & 91.0\% \\
DivPrune~{\scriptsize(CVPR25)} & 75.6 & 60.0 & 71.4 & 56.3 & 81.9 & 1436.7 & 65.1 & \textbf{60.9} & 67.9 & 94.4\% \\
\rowcolor{subsetbg}
\textbf{\method{} (Ours)} & \textbf{77.8} & \textbf{62.2} & \textbf{71.7} & 56.7 & \textbf{88.3} & \textbf{1476.9} & \textbf{65.9} & 60.1 & \textbf{69.6} & \textbf{96.8\%} \\
\bottomrule
\end{tabular}%
}
\end{table*}

\clearpage
\section{Additional Qualitative Visualizations}
\label{app:qualitative_visualizations}
\begin{figure*}[h!]
    \centering
    \resizebox{!}{0.76\textheight}{%
        \includegraphics{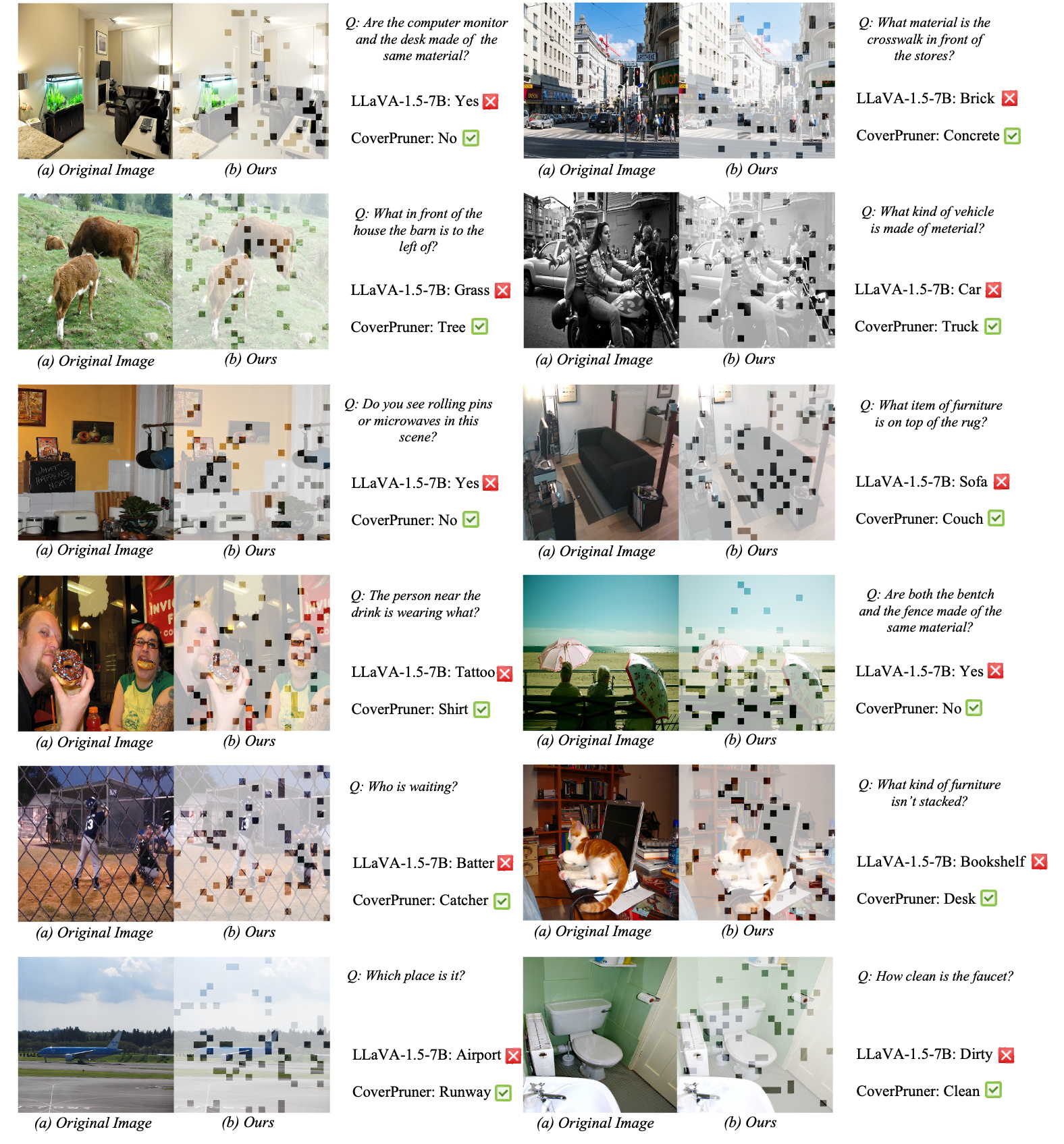}%
    }
    \caption{\textbf{Examples from the GQA dataset where \method{} answers correctly under aggressive pruning while the full LLaVA-1.5-7B model fails, covering routine and relational questions.} Check and cross marks indicate correct and incorrect responses.}
    \label{fig:app_cp_wins}
\end{figure*}

\begin{figure*}[h!]
    \centering
    \resizebox{!}{0.72\textheight}{%
        \includegraphics{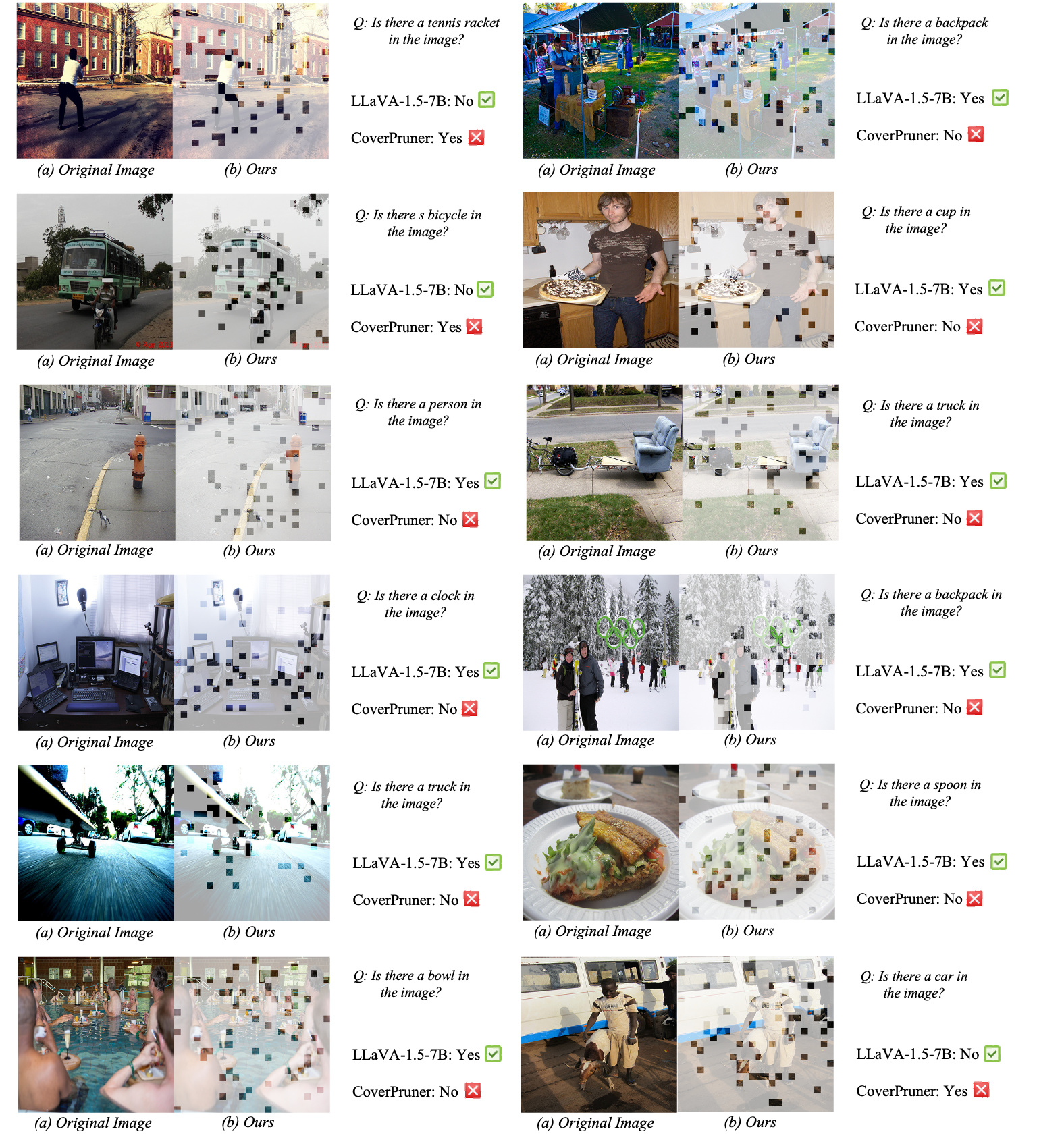}%
    }
    \caption{\textbf{Examples from the POPE dataset where the full LLaVA-1.5-7B model answers correctly while \method{} fails under aggressive pruning.} Check and cross marks indicate correct and incorrect responses.}
    \label{fig:app_llava_wins}
\end{figure*}

\begin{figure*}[h!]
    \centering
    \resizebox{!}{0.76\textheight}{%
        \includegraphics{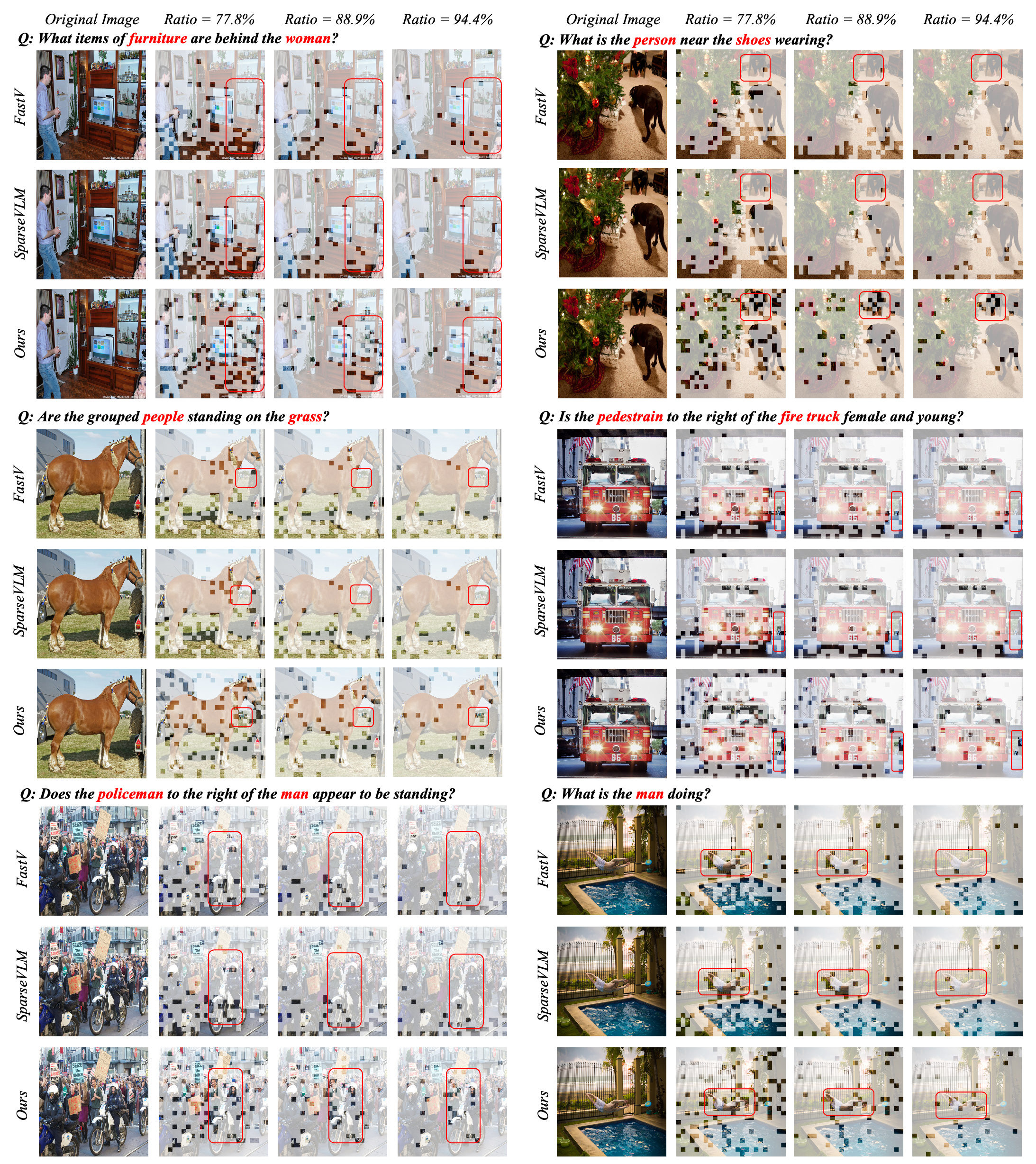}%
    }
    \caption{\textbf{Qualitative comparison of different pruning methods and our proposed method CoverPruner on the GQA dataset.} The figure presents original images alongside their pruned versions at pruning ratios of 77.8\%, 88.9\%, and 94.4\%. Bounding boxes highlight the key semantic regions aligned with the text query. CoverPruner better preserves these regions, especially under aggressive pruning ratios (e.g., 94.4\%).}
    \label{fig:app_pruning_comparison}
\end{figure*}

This section provides additional qualitative case studies and pruning visualizations. Figure~\ref{fig:app_cp_wins} and Figure~\ref{fig:app_llava_wins} compare \method{} with the full LLaVA-1.5-7B model, and Figure~\ref{fig:app_pruning_comparison} further visualizes retained visual tokens across different pruning methods.

Figures~\ref{fig:app_cp_wins} and~\ref{fig:app_llava_wins} show the two sides of coverage-based pruning. In Figure~\ref{fig:app_cp_wins}, \method{} answers correctly while the full LLaVA-1.5-7B model fails. These are mostly questions that depend on the broader scene rather than a single salient object, and the coverage objective helps by making sure every discarded token still has a representative in the retained set, so the contextual evidence is not lost even at aggressive budgets. Figure~\ref{fig:app_llava_wins} shows the opposite case on the POPE dataset. When the queried object is very small or visually unclear in the image, even the retained tokens cannot capture enough detail for the model to recognize it correctly. For example, the model may mistake a bus for a car, which leads to wrong existence judgments. This matches the limitation discussed in Section~\ref{sec:limitations}: coverage works best when redundancy dominates, and is weaker when the answer depends on fine-grained details of small or ambiguous objects.

Figure~\ref{fig:app_pruning_comparison} compares retained-token patterns across FastV, SparseVLM, and \method{} on LLaVA-1.5-7B at three pruning ratios (77.8\%, 88.9\%, and 94.4\%), further demonstrating the effectiveness of \method{}. Two patterns stand out. \method{} spreads retained tokens according to local content: fewer tokens are kept in repetitive regions such as sky, grass, or road, and more tokens are kept where the scene is visually diverse, such as people, vehicles, and signs. In contrast, FastV and SparseVLM concentrate tokens around a few salient anchors and leave large parts of the image unrepresented, which is most visible at the 94.4\% budget. \method{} is also query-aware: guided by the question, it places more retained tokens on the objects mentioned in the query while still keeping enough tokens on the rest of the scene, so the model can both answer the specific question and understand the surrounding context that relational questions depend on.

\clearpage
\section{Additional Ablations}
\label{sec:appendix_ablation}
This appendix provides additional ablations that complement the main results in Section~\ref{sec:ablation}. Except for the runtime breakdown, experiments use LLaVA-1.5-7B with 64 retained tokens and report TextVQA and POPE where applicable. Unless a component is explicitly ablated, the configuration is mean-centered similarity, first-layer attention relevance with $\tau\!=\!0.3$, and standard sum aggregation, matching Section~\ref{sec:ablation}.

\subsection{Runtime Overhead Breakdown}
\label{app:overhead_breakdown}

Table~\ref{tab:app_overhead} breaks down the pruning overhead of \method{} into pairwise similarity computation, first-layer attention probing, and demand-weighted coverage selection. Similarity and attention probing are computed once over the original visual-token sequence, so they are shared across budgets for the same model; the iterative coverage-selection step increases with the retained budget. The LLaVA-NeXT rows use the 2,880-token source sequence, with the 320-token row matching Table~\ref{tab:efficiency}.

\begin{table}[h!]
\caption{\textbf{Overhead breakdown of \method{} components.} Values are pruning-overhead ms/example before decoder prefill, while Table~\ref{tab:efficiency} reports decoder latency in ms/token.}
\label{tab:app_overhead}
\centering
\scriptsize
\setlength{\tabcolsep}{2.5pt}
\renewcommand{\arraystretch}{0.9}
\resizebox{\columnwidth}{!}{%
\begin{tabular}{lccccc}
\toprule
Model & Budget & Sim. & Attn. & Coverage Sel. & Total \\
 &  & \multicolumn{4}{c}{(ms/example)} \\
\midrule
\multirow{3}{*}{LLaVA-1.5} & 128 & \multirow{3}{*}{0.052} & \multirow{3}{*}{0.031} & 0.092 & 0.175 \\
 & 64  & & & 0.060 & 0.143 \\
 & 32  & & & 0.038 & 0.121 \\
\midrule
\multirow{3}{*}{LLaVA-NeXT-7B} & 640 & \multirow{3}{*}{0.940} & \multirow{3}{*}{0.190} & 0.742 & 1.872 \\
 & 320 & & & 0.391 & 1.521 \\
 & 160 & & & 0.233 & 1.363 \\
\bottomrule
\end{tabular}%
}
\renewcommand{\arraystretch}{1.0}
\vspace{-2mm}
\end{table}

\subsection{Mean-Centering of the Similarity Matrix}
\label{app:meancenter}

Section~\ref{sec:query} motivated mean-centering as a calibration that gives the $\max$-based coverage usable dynamic range when raw cosine similarities concentrate in a narrow band. With LLM-embed relevance at $\tau\!=\!0.3$, mean-centering improves TextVQA from 54.79 to 55.23. The gain is small but consistent with the main configuration, where the deployed objective uses mean-centered non-negative utilities.

\subsection{Query-Conditioning Temperature}
\label{app:temperature}

Table~\ref{tab:app_tau} compares unweighted coverage ($\tau\!=\!0$) against our default ($\tau\!=\!0.3$) with first-layer attention relevance and mean-centered similarity. The gain is modest but consistent on both benchmarks, and \method{} is robust to the exact value of $\tau$ in this range; combined with the relevance-signal ordering in the main ablation, this supports our claim that the quality of the relevance signal matters more than the strength of conditioning.

\begin{table}[h!]
\caption{\textbf{Effect of query-conditioning temperature} $\tau$ on LLaVA-1.5-7B, 64 retained tokens, with first-layer attention relevance and mean-centered similarity.}
\label{tab:app_tau}
\centering
\setlength{\tabcolsep}{6pt}
\begin{tabular}{lcc}
\toprule
Configuration & TextVQA & POPE \\
\midrule
$\tau\!=\!0$ (unweighted) & 54.93 & 85.5 \\
$\tau\!=\!0.3$ (default) & 55.62 & 86.1 \\
\bottomrule
\end{tabular}
\end{table}

\subsection{Representation Space for Coverage Similarity}
\label{app:repspace}

Section~\ref{sec:query} motivated computing coverage similarity in the projector output space rather than in the vision-encoder's native feature space, since the projector output is the representation actually consumed by the LLM decoder. Table~\ref{tab:app_repspace} isolates this choice: we hold the RCM objective, demand weights, and greedy selection protocol fixed, and change only the features used to compute cosine similarity in Eq.~\ref{eq:cov_single}, on LLaVA-1.5-7B with 64 retained tokens.

Using pre-projector encoder features in place of projector-output features causes a substantial drop on both benchmarks. This supports our design choice of measuring redundancy and representativeness in the space where visual evidence actually enters the language model, rather than in the upstream vision-encoder space: two patches can be near-duplicates to the vision encoder while providing distinguishable evidence to the LLM after projection, or vice versa.

\begin{table}[h!]
\caption{\textbf{Effect of similarity representation space} on LLaVA-1.5-7B, 64 retained tokens, with the RCM objective and selection protocol otherwise unchanged.}
\label{tab:app_repspace}
\centering
\setlength{\tabcolsep}{6pt}
\resizebox{0.45\textwidth}{!}{%
\begin{tabular}{lcc}
\toprule
Similarity space & TextVQA & POPE \\
\midrule
Encoder (pre-projector) & 48.3 & 81.5 \\
Projector output (default) & \textbf{55.6} & \textbf{86.1} \\
\midrule
Absolute gain & \textbf{+7.3} & \textbf{+4.6} \\
\bottomrule
\end{tabular}
}
\end{table}

\subsection{Variants Considered}
\label{app:variants}
We also considered two alternatives to the sum aggregation in Eq.~\ref{eq:coverage_step}: a Top-$K$ mean over the largest residual gains, and a concave-over-modular reformulation. Table~\ref{tab:app_variants} reports their behaviour. Top-$K$ mean ($K\!=\!20$) targets the density bias of the sum aggregation by capping the number of demand-side contributors that any single candidate can accumulate gain from. It marginally outperforms the unweighted sum baseline on TextVQA but does not improve POPE, indicating that density bias is task-dependent and most pronounced on OCR-heavy benchmarks. We therefore do not deploy it by default.

The concave-over-modular form, $f(S)\!=\!\sum_i w_i \sqrt{\sum_{j \in S} \mathrm{sim}(\mathbf{v}_i, \mathbf{v}_j)}$, replaces the $\max$ in Eq.~\ref{eq:weighted_coverage} with a concave function of the cumulative similarity to $S$. Despite being a natural alternative, it substantially underperforms in practice. The reason is that the $\sum$-based accumulation collapses the ``best representative'' semantics that the $\max$ operator provides: a token whose similarity to $S$ comes from many distant retained tokens receives the same coverage as one served by a single nearby retained token, which is exactly the opposite of what pruning needs. We report this as a negative result.

\begin{table}[h!]
\caption{\textbf{Aggregation variants considered} but not deployed (LLaVA-1.5-7B, 64 retained tokens). Default is sum aggregation with $\tau\!=\!0.3$.}
\label{tab:app_variants}
\centering
\small
\setlength{\tabcolsep}{6pt}
\resizebox{0.48\textwidth}{!}{%
\begin{tabular}{lcc}
\toprule
Variant & TextVQA & POPE \\
\midrule
Sum aggregation, $\tau\!=\!0$ (unweighted) & 54.93 & 85.5 \\
Top-$K$ Mean ($K\!=\!20$, $\tau\!=\!0$) & 55.36 & 85.5 \\
Concave-over-modular ($\sqrt{\cdot}$) & 44.97 & 64.6 \\
\midrule
Sum aggregation, $\tau\!=\!0.3$ (default) & 55.62 & 86.1 \\
\bottomrule
\end{tabular}
}
\end{table}

\end{document}